\documentclass[letterpaper]{article} 
\usepackage{aaai2027}
\usepackage[hyphens]{url}  
\usepackage{graphicx} 
\usepackage{natbib}  
\usepackage{caption} 
\usepackage{algorithm}
\usepackage{booktabs}
\usepackage{amsmath}
\usepackage{amssymb}
\usepackage{algpseudocode}
\usepackage{graphicx}
\usepackage{multirow}

\newcommand{\best}[1]{\textbf{#1}}
\newcommand{\second}[1]{\underline{#1}}

\title{RippleKV: Cross-Layer KV Cache Allocation via Perturbation Propagation}

\author{
    Dongjie Xu\textsuperscript{\rm 1},
    Kai Qian\textsuperscript{\rm 1},
    Julius\textsuperscript{\rm 2},
    Weijie Shi\textsuperscript{\rm 3},
    Yuxuan Sun\textsuperscript{\rm 4},\\
    Minghua Tang\textsuperscript{\rm 5},
    Fenglei Jin\textsuperscript{\rm 2},
    Hanchi Dong\textsuperscript{\rm 4},
    Jiajie Xu\textsuperscript{\rm 1}
}
\affiliations{
    \textsuperscript{\rm 1}Soochow University\\
    \textsuperscript{\rm 2}Tencent Inc.\\
    \textsuperscript{\rm 3}The Hong Kong University of Science and Technology\\
    \textsuperscript{\rm 4}Beijing University of Posts and Telecommunications\\
    \textsuperscript{\rm 5}Jiangnan University
}

\begin{document}

\maketitle

\begin{abstract}
Long-context LLM inference is bottlenecked by KV cache memory, yet distributing a limited cache budget across layers remains challenging. Existing methods rely on proxies such as layer depth, attention statistics, or representation change. These proxies do not measure how perturbations at each layer propagate to the output and may therefore cause sensitive layers to be underallocated while tolerant layers are overallocated. To address this issue, we propose RippleKV, which allocates cache across layers by estimating how perturbations to each layer's value cache affect the final predictive distribution. RippleKV independently injects norm-adaptive perturbations into each layer's value cache and measures the induced KL divergence at the model output over a small calibration set. Averaging these responses yields a sensitivity profile specific to the model that need not vary monotonically with depth. RippleKV then converts the sensitivity profile into layer budget multipliers by normalizing the sensitivity scores and applying an exponential mapping. A ratio parameter controls the allocation disparity between sensitive and tolerant layers, while a final normalization preserves the KV cache budget. Experiments on LongBench demonstrate that RippleKV achieves the highest average performance among the evaluated KV cache compression methods under matched cache budgets.

\end{abstract}


\section{Introduction}

Large language models (LLMs) have been increasingly extended to long-context scenarios, enabling applications such as document summarization, question answering, information retrieval, and code generation~\citep{roziere2023code,shaham2023zeroscrolls}. These applications often require processing long documents or multi-hop evidence~\citep{zhang2024benchmarking,kamalloo2023evaluating}. 
However, during autoregressive decoding, the KV cache grows linearly with the context length and the number of Transformer layers, leading to substantial memory and bandwidth pressure~\citep{xiao2024efficient,zhang2023h2o,liu2024kivi,adnan2024keyformer}. Therefore, KV cache compression has become critical for efficient long-context LLM inference.


Existing KV cache compression methods reduce memory usage by retaining only a subset of cached tokens. StreamingLLM~\citep{xiao2024efficient} preserves attention sinks and recent tokens, while Scissorhands, H$_2$O, and SnapKV~\citep{liu2023scissorhands,zhang2023h2o,li2024snapkv} select important tokens based on persistent or accumulated attention scores. More recent layer-aware methods further assign different cache budgets across Transformer layers to improve cache utilization~\citep{cai2024pyramidkv,DBLP:conf/iclr/QinCLHFCLL25}. These methods typically determine layer budgets using proxy signals such as layer depth, attention statistics, or representation changes. However, these proxies do not directly quantify the relative cache requirements of different layers under a fixed global budget, potentially underallocating sensitive layers while overallocating tolerant ones.


A principled allocation strategy should capture not only the structural or local characteristics of each layer, but also how strongly changes to its cache propagate to the final model output. Our analysis reveals that cache requirements vary substantially across layers and exhibit a nonmonotonic relationship with depth, with layers that incur greater damage under compression often appearing at intermediate positions. This heterogeneity cannot be adequately captured by fixed depth schedules or layer-local statistics alone. More importantly, we find that the shift in the final predictive distribution caused by perturbing a layer's Value cache is strongly aligned with the actual damage caused by compressing that layer. These observations suggest that downstream output sensitivity provides a more faithful basis for distributing a fixed global cache budget across layers.


Motivated by these observations, we propose RippleKV, a cross-layer KV cache allocation method guided by downstream output sensitivity. During offline profiling, RippleKV perturbs the Value cache of one layer at a time while keeping its Key cache and all other layers unchanged, and measures the induced KL divergence in the final output distribution. Averaging these responses produces a model specific sensitivity profile that reflects how strongly cache changes at each layer propagate to the model output. RippleKV then converts this profile into layer budgets through a scale invariant exponential mapping, where a ratio parameter controls the allocation disparity across layers while normalization preserves the global cache budget. Layers with stronger downstream effects receive larger budgets, whereas more tolerant layers are compressed more aggressively. Since RippleKV only redistributes cache capacity across layers while retaining the original token scoring and selection strategy, it can be integrated with existing KV cache compression methods without additional model evaluations during inference. Experiments on LongBench across three model families and multiple cache budgets show that RippleKV achieves the highest average performance in all evaluated settings while maintaining comparable runtime and memory efficiency.

Our contributions are summarized as follows:
\begin{itemize}
    \item We demonstrate that layer-wise compression damage is highly heterogeneous and nonmonotonic with depth, and that the final output response to controlled Value cache perturbations provides a more faithful signal for cross-layer budget allocation than layer position.
    \item We introduce RippleKV, a cross-layer KV cache allocation framework built on perturbation propagation. RippleKV estimates layer sensitivity from the effect of controlled Value cache perturbations on the final output and derives model specific budgets through a scale invariant allocation rule under a fixed global cache constraint.
    \item Extensive experiments on LongBench show that RippleKV achieves the best average performance across three model families and multiple cache budgets while maintaining comparable runtime and memory efficiency.
\end{itemize}

\section{Related Work}


\paragraph{KV Cache Compression.} KV cache compression reduces the memory and computational overhead of long-context inference by retaining only important tokens. StreamingLLM~\cite{xiao2024efficient} preserves attention sinks and recent tokens for stable streaming generation, while H$_2$O~\cite{zhang2023h2o} retains recent tokens and heavy hitters identified by accumulated attention scores. TOVA~\cite{oren2024transformers} performs online eviction by removing the least-attended token at each decoding step. SnapKV~\cite{li2024snapkv} estimates token importance from an observation window at the end of the prompt. These methods primarily determine which tokens to retain within each layer, while generally using uniform or predefined cache capacities across layers.


\paragraph{Layer-Wise KV Cache Allocation.} Recent studies have shown that different Transformer layers have heterogeneous cache requirements, motivating non-uniform allocation under a fixed global cache budget. PyramidKV~\cite{cai2024pyramidkv} and PyramidInfer~\cite{yang2024pyramidinfer} allocate progressively smaller cache budgets to higher layers based on pyramidal information patterns. More recent methods dynamically determine layer budgets using task-dependent attention patterns, spatial and temporal attention dynamics, Key representation deviation, or intermediate reconstruction errors~\cite{zhou2024dynamickv,DBLP:conf/iclr/QinCLHFCLL25,DBLP:journals/corr/abs-2509-17396,DBLP:conf/emnlp/ShenYZWJN25,lin2025compresskv}. Existing approaches therefore mainly rely on structural priors, attention statistics, or layer-local discrepancies, which may not fully capture how compression effects propagate through subsequent layers. In contrast, RippleKV allocates layer-wise cache budgets based on how cache perturbations affect the final output distribution.

\section{Rethinking Cache Allocation Across Layers}
\label{sec:motivation}

\subsection{Compression Damage Does Not Follow a Fixed Depth Pattern}
\label{sec:layer-sensitivity}

Existing layer-wise allocation strategies often assign cache budgets according to layer depth. However, layer position does not directly quantify how compressing a layer's KV cache affects the final model prediction. We therefore conduct an isolated single-layer compression analysis to obtain an empirical measure of layer sensitivity. For each layer $\ell$, we compress only its KV cache while keeping all other layers uncompressed, using the same compression setting for every layer. Let $p_i^{\mathrm{full}}$ denote the predictive distribution of the model with the full cache for input $x_i$, and let $p_{i,\ell}^{\mathrm{comp}}$ denote the corresponding distribution when only layer $\ell$ is compressed. We define the compression damage of layer $\ell$ as
\begin{equation}
D_\ell =
\frac{1}{N}
\sum_{i=1}^{N}
D_{\mathrm{KL}}
\left(
p_i^{\mathrm{full}}
\parallel
p_{i,\ell}^{\mathrm{comp}}
\right).
\label{eq:layer-compression-damage}
\end{equation}
A larger $D_\ell$ indicates that compressing layer $\ell$ causes a greater change in the predictive distribution and should therefore be treated more conservatively during cache allocation.

Across the 32 layers of Llama-3.1-8B-Instruct, the mean isolated compression damage spans $0.0030$--$0.0524$, a $17.7\times$ difference, with the maximum occurring at an intermediate layer. Layer index correlates only weakly with compression damage (mean $|\rho|=0.359$; Table~\ref{tab:sensitivity-correlation}), showing that depth captures only a coarse trend and can lead fixed schedules to underallocate cache to sensitive layers while overallocating it to tolerant ones.

\begin{figure*}[t]
    \centering
    \includegraphics[width=0.95\textwidth]{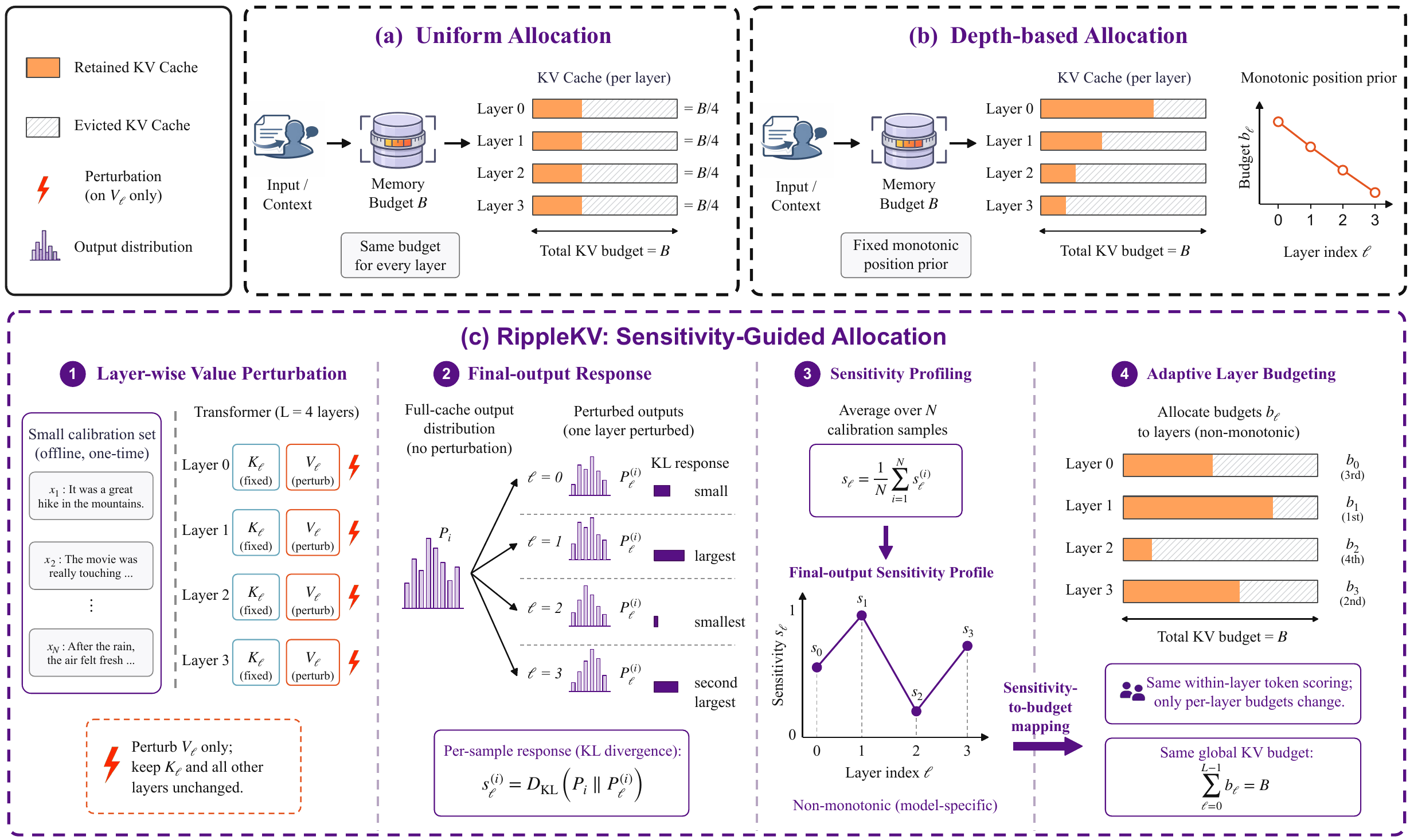}
    \caption{Overview of RippleKV. Unlike uniform or depth based allocation, RippleKV measures the final output response to layer specific Value cache perturbations and allocates the global cache budget according to the resulting sensitivity profile.}
    \label{fig:overview}
\end{figure*}

\subsection{Perturbation Response Captures Compression Sensitivity}
\label{sec:perturbation-response}

Although $D_\ell$ directly measures compression damage, obtaining it requires applying a specific compression operation to every layer, making it unsuitable as a general allocation signal. We instead perturb the value cache of one layer at a time while leaving all other layers unchanged and measure the resulting shift in the final predictive distribution. Let $p_{i,\ell}^{\mathrm{pert}}$ denote the output distribution after perturbing layer $\ell$. We define the perturbation response as
\begin{equation}
s_\ell =
\frac{1}{N}
\sum_{i=1}^{N}
D_{\mathrm{KL}}
\left(
p_i^{\mathrm{full}}
\parallel
p_{i,\ell}^{\mathrm{pert}}
\right).
\label{eq:perturbation-response}
\end{equation}
Because $s_\ell$ is measured at the final output after the perturbation propagates through subsequent layers, it captures the end-to-end effect of modifying a layer's cache rather than its immediate local effect~\citep{jing2026beyond,dong2020hawq}.

To assess ranking quality, we compute the absolute Spearman correlation between each signal and the isolated compression damage in Eq.~\eqref{eq:layer-compression-damage}.

\begin{table}[H]
\centering
\small
\setlength{\tabcolsep}{7pt}
\begin{tabular}{lcc}
\toprule
Layer signal & Mean $|\rho|$ $\uparrow$ & Range \\
\midrule
Layer index
& 0.359 & 0.218--0.480 \\
RippleKV perturbation response
& \textbf{0.799} & \textbf{0.723--0.857} \\
\bottomrule
\end{tabular}
\caption{Absolute Spearman correlation with isolated compression damage.}
\label{tab:sensitivity-correlation}
\end{table}

As shown in Table~\ref{tab:sensitivity-correlation}, the RippleKV perturbation response achieves a mean correlation of $0.799$, substantially exceeding layer index ($0.359$). Its correlation remains between $0.723$ and $0.857$, showing that the final-output response consistently captures compression sensitivity beyond the coarse depth prior. This result motivates its use for layer-wise cache allocation.

\section{Methodology}
\label{sec:method}

\subsection{Problem Formulation}
\label{sec:formulation}

Given a decoder only Transformer with $L$ layers and an input sequence
$\mathbf{x}=\{x_t\}_{t=1}^{T}$, the prefilling stage produces a KV cache at each layer:
\begin{equation}
\mathcal{C}_{\ell}
=
\{(\mathbf{k}_{\ell,t},\mathbf{v}_{\ell,t})\}_{t=1}^{T},
\qquad
\ell=0,\ldots,L-1,
\end{equation}
where $\mathbf{k}_{\ell,t}$ and $\mathbf{v}_{\ell,t}$ denote the key and value states of token $t$ at layer $\ell$. During autoregressive decoding, each new query attends to the cached states of previous tokens. The total KV cache size therefore grows linearly with both the sequence length and the number of layers, while the attention cost increases with the number of retained tokens.

KV cache compression reduces these costs by retaining only a subset of token positions at each layer. Let
$\mathcal{S}_{\ell}\subseteq\{1,\ldots,T\}$ denote the positions retained at layer $\ell$. The resulting compressed cache is
\begin{equation}
\widehat{\mathcal{C}}_{\ell}
=
\{(\mathbf{k}_{\ell,t},\mathbf{v}_{\ell,t})
\mid t\in\mathcal{S}_{\ell}\},
\end{equation}
where $b_\ell=|\mathcal{S}_{\ell}|$ is the cache budget assigned to layer $\ell$. Under a global cache budget $B$, the layer budgets satisfy
\begin{equation}
\sum_{\ell=0}^{L-1} b_\ell = B.
\label{eq:global-budget}
\end{equation}

Given a fixed token selection strategy, we seek a layer budget allocation $\mathbf{b}=\{b_\ell\}_{\ell=0}^{L-1}$ that preserves generation quality under the global budget constraint in Eq.~\eqref{eq:global-budget}.

\subsection{Overview of RippleKV}
\label{sec:method-overview}

As illustrated in Figure~\ref{fig:overview}, RippleKV consists of offline sensitivity profiling followed by cache compression during inference. Given a small calibration set, it first obtains the reference output distributions using the full cache. RippleKV then perturbs the Value cache of one compressible layer at a time, while keeping its Key cache and all other caches unchanged. The resulting change in the final output distribution is measured by KL divergence and averaged across calibration examples to form a sensitivity profile specific to the model. Since the response is measured after propagating through the subsequent layers, it captures the influence of each layer on the final prediction.

The sensitivity profile is then converted into layer budgets
$\mathbf{b}=\{b_\ell\}_{\ell\in\mathcal{A}}$
under the global constraint
$\sum_{\ell\in\mathcal{A}}b_\ell=B$.
Layers with stronger responses receive larger budgets, while less sensitive layers are compressed more aggressively. This produces a nonmonotonic allocation while preserving the global cache budget. During inference, RippleKV changes only the budget assigned to each layer and retains the original token scoring strategy. The sensitivity profile is computed once offline and reused without additional model evaluations. The complete procedure is summarized in Algorithm~\ref{alg:ripplekv}.

\begin{algorithm}[t]
\caption{RippleKV}
\label{alg:ripplekv}
\begin{algorithmic}[1]

\Statex \textbf{Input:} Model $\mathcal{M}$; calibration set
$\mathcal{D}_{\mathrm{cal}}=\{\mathbf{x}_i\}_{i=1}^{N}$;
compressible layers $\mathcal{A}$; global cache budget $B$;
perturbation strength $\alpha$; target budget ratio $r$;
weight bound $\gamma$

\Statex \textbf{Output:} Sensitivity profile $\mathbf{s}$
and layer budgets $\mathbf{b}$

\Statex \textbf{Sensitivity Profiling}
\State $s_\ell\gets 0,\quad\forall\ell\in\mathcal{A}$

\ForAll{$\mathbf{x}_i\in\mathcal{D}_{\mathrm{cal}}$}
    \State $\mathbf{P}_i
    \gets
    \Call{ReferencePredict}{\mathcal{M},\mathbf{x}_i}$

    \ForAll{$\ell\in\mathcal{A}$}
        \State $\mathbf{Q}_{i,\ell}
        \gets
        \Call{PerturbedPredict}
        {\mathcal{M},\mathbf{x}_i,\ell,\alpha}$

        \State $d_{i,\ell}
        \gets
        \Call{WindowKL}
        {\mathbf{P}_i,\mathbf{Q}_{i,\ell}}$

        \State $s_\ell\gets s_\ell+d_{i,\ell}$
    \EndFor
\EndFor

\State $s_\ell\gets s_\ell/N,
\quad\forall\ell\in\mathcal{A}$

\Statex \textbf{Budget Allocation}
\State $s_{\min}\gets\min_{\ell\in\mathcal{A}}s_\ell$,
$s_{\max}\gets\max_{\ell\in\mathcal{A}}s_\ell$

\ForAll{$\ell\in\mathcal{A}$}
    \State $\widehat{s}_\ell
    \gets
    \dfrac{s_\ell-s_{\min}}
    {s_{\max}-s_{\min}+\epsilon}$

    \State $\widetilde{w}_\ell
    \gets
    r^{\widehat{s}_\ell}$
\EndFor

\State $\mathbf{w}
\gets
\Call{ClipNormalize}
{\widetilde{\mathbf{w}},\gamma}$

\ForAll{$\ell\in\mathcal{A}$}
    \State $b_\ell
    \gets
    \operatorname{round}
    \left(
    \dfrac{B}{|\mathcal{A}|}w_\ell
    \right)$
\EndFor

\State \Return $\mathbf{s},\mathbf{b}$

\end{algorithmic}
\end{algorithm}

\subsection{Layer Sensitivity Estimation}
\label{sec:sensitivity-estimation}

Given a calibration set
$\mathcal{D}_{\mathrm{cal}}=\{\mathbf{x}_i\}_{i=1}^{N}$,
we divide each example into a prefix used to construct the KV cache and an evaluation window $\mathcal{W}_i$. We first run the model with the intact cache and record the reference output distribution
$\mathbf{P}_{i,t}$ at each position
$t\in\mathcal{W}_i$. For every compressible layer $\ell$, we then apply a controlled perturbation to its prefix Value cache and evaluate the same output window. The intervention is restricted to layer $\ell$: its Key cache and the caches of all other layers remain unchanged, and no cached position is removed during profiling.

Let $\mathbf{v}_{i,\ell,h,u}$ denote the Value vector at layer $\ell$, KV head $h$, and prefix position $u$. For each position eligible for compression, RippleKV constructs
\begin{equation}
\widetilde{\mathbf{v}}_{i,\ell,h,u}
=
\mathbf{v}_{i,\ell,h,u}
+
\alpha
\left\|\mathbf{v}_{i,\ell,h,u}\right\|_2
\boldsymbol{\epsilon}_{i,\ell,h,u},
\label{eq:value-perturbation}
\end{equation}
where
$\boldsymbol{\epsilon}_{i,\ell,h,u}
\sim\mathcal{N}(\mathbf{0},\mathbf{I})$
and $\alpha$ controls the perturbation strength. Scaling the noise by the norm of each Value vector adapts the perturbation to its local magnitude and reduces sensitivity to absolute activation scale differences across layers and tokens. Positions protected by the base compression method are excluded from perturbation. Since the Key cache is fixed, this intervention leaves the attention weights at the target layer unchanged while modifying the content aggregated from its cached values.

Let $\mathbf{Q}_{i,\ell,t}$ denote the output distribution at position $t$ after perturbing layer $\ell$. We measure the response of layer $\ell$ on example $\mathbf{x}_i$ by averaging the output divergence over the evaluation window:
\begin{equation}
d_{i,\ell}
=
\frac{1}{|\mathcal{W}_i|}
\sum_{t\in\mathcal{W}_i}
D_{\mathrm{KL}}
\left(
\mathbf{P}_{i,t}
\parallel
\mathbf{Q}_{i,\ell,t}
\right).
\label{eq:sample-response}
\end{equation}
The sensitivity estimate for layer $\ell$ is then obtained by averaging across the calibration set:
\begin{equation}
s_\ell
=
\frac{1}{N}
\sum_{i=1}^{N}d_{i,\ell}.
\label{eq:layer-sensitivity}
\end{equation}
A larger $s_\ell$ indicates that the model output is more responsive to a controlled Value cache perturbation at layer $\ell$. Repeating the procedure for all compressible layers yields the model specific sensitivity profile
$\mathbf{s}=\{s_\ell\}_{\ell\in\mathcal{A}}$.

The profiling procedure requires no gradient computation and is performed only once for each model. The resulting sensitivity profile is independent of the target cache budget and can therefore be reused across different compression settings. During inference, RippleKV directly applies the corresponding layer budgets without additional perturbation evaluations.

\begin{table*}[t]
\centering

\setlength{\tabcolsep}{2.8pt}
\renewcommand{\arraystretch}{1.05}
\resizebox{\textwidth}{!}{%
\begin{tabular}{lccccccccccccccccc}
\toprule
\multirow{2}{*}{Method}
& \multicolumn{3}{c}{Single-Document QA}
& \multicolumn{3}{c}{Multi-Document QA}
& \multicolumn{3}{c}{Summarization}
& \multicolumn{3}{c}{Few-shot Learning}
& \multicolumn{2}{c}{Synthetic}
& \multicolumn{2}{c}{Code}
& \multirow{2}{*}{Avg.} \\
\cmidrule(lr){2-4}
\cmidrule(lr){5-7}
\cmidrule(lr){8-10}
\cmidrule(lr){11-13}
\cmidrule(lr){14-15}
\cmidrule(lr){16-17}
& NrtvQA & Qasper & MF-en
& HotpotQA & 2WikiMQ & Musique
& GovReport & QMSum & MultiNews
& TREC & TriviaQA & SAMSum
& PCount & PRe
& Lcc & RB-P
& \\
\midrule

\multicolumn{18}{c}{Llama-3.1-8B-Instruct, Full Cache} \\
\midrule
Full Cache
& 30.72 & 47.06 & 55.28
& 59.51 & 51.83 & 32.66
& 35.25 & 24.94 & 27.05
& 29.50 & 91.71 & 40.92
& 10.70 & 100.00
& 54.13 & 47.58
& 46.18 \\
\midrule

\multicolumn{18}{c}{Llama-3.1-8B-Instruct, KV Cache Budget = 10\%} \\
\midrule
StreamingLLM
& 21.67 & 18.44 & 23.03
& 37.24 & 22.28 & 14.06
& 25.03 & 19.13 & 20.05
& 28.00 & 90.70 & 35.31
& 4.00 & 16.00
& \second{52.35} & \best{52.57}
& 29.99 \\

H$_2$O
& 15.72 & \best{28.97} & 20.19
& 32.81 & \best{28.04} & 10.39
& \best{27.92} & \best{21.25} & \best{23.56}
& \best{39.00} & 89.35 & 34.80
& \best{10.42} & 54.00
& 50.17 & 46.08
& 33.29 \\

SnapKV
& \best{24.33} & 20.80 & \second{23.63}
& \best{44.74} & 24.10 & \second{19.88}
& \second{25.56} & 19.80 & 20.08
& 33.50 & \second{91.49} & \second{39.84}
& 5.50 & \second{54.50}
& 51.09 & \second{48.94}
& \second{34.24} \\

PyramidKV
& 23.39 & 20.76 & 23.30
& \second{44.71} & 24.78 & 18.66
& 24.98 & \second{20.68} & 19.96
& 33.50 & \best{91.66} & 39.59
& \second{6.00} & \second{54.50}
& 51.06 & 48.63
& 34.13 \\

\textbf{RippleKV}
& \second{23.62} & \second{22.54} & \best{24.22}
& 44.37 & \second{27.84} & \best{20.46}
& \second{25.56} & 20.35 & \second{20.40}
& \second{34.00} & \second{91.49} & \best{40.62}
& \second{6.00} & \best{58.50}
& \best{52.59} & 48.57
& \best{35.07} \\
\midrule

\multicolumn{18}{c}{Llama-3.1-8B-Instruct, KV Cache Budget = 20\%} \\
\midrule
StreamingLLM
& 22.18 & 23.09 & 24.92
& 41.44 & 27.23 & 18.16
& 28.08 & 20.14 & 22.68
& 28.00 & \second{91.71} & 34.94
& 6.00 & 28.50
& 52.18 & \best{51.16}
& 32.53 \\

H$_2$O
& 20.36 & \best{33.74} & 27.47
& 39.07 & 35.45 & 16.34
& \best{30.43} & \best{22.93} & \best{25.56}
& \best{39.00} & 89.83 & 36.91
& \best{9.54} & 76.50
& 50.78 & \second{47.81}
& 37.61 \\

SnapKV
& \best{28.57} & 25.39 & \second{31.19}
& \second{54.40} & 38.20 & \best{25.35}
& 27.92 & 21.70 & \second{22.75}
& \second{35.00} & \best{92.31} & 41.13
& 8.16 & \second{86.50}
& 53.42 & 46.92
& \second{39.93} \\

PyramidKV
& \second{27.16} & 25.53 & 29.92
& \best{54.64} & \best{40.51} & 24.29
& 27.34 & 21.65 & 22.74
& 32.00 & 91.56 & \second{41.33}
& \second{8.56} & \second{86.50}
& \second{53.53} & 47.32
& 39.66 \\

\textbf{RippleKV}
& 26.50 & \second{27.88} & \best{32.87}
& 52.68 & \second{38.80} & \second{24.85}
& \second{28.24} & \second{21.98} & \second{22.75}
& \second{35.00} & 91.38 & \best{42.22}
& 8.10 & \best{90.00}
& \best{54.52} & \second{47.81}
& \best{40.35} \\
\midrule

\multicolumn{18}{c}{Llama-3.1-8B-Instruct, KV Cache Budget = 30\%} \\
\midrule
StreamingLLM
& 24.23 & 28.29 & 26.57
& 43.67 & 30.92 & 21.10
& 29.32 & 21.02 & \second{24.23}
& 31.50 & \best{91.49} & 36.79
& 6.25 & 36.50
& 50.90 & \best{49.86}
& 34.54 \\

H$_2$O
& 20.20 & \best{37.40} & 34.39
& 45.50 & 42.38 & 19.83
& \best{31.69} & \best{23.26} & \best{25.67}
& \best{40.00} & 90.99 & 37.92
& 9.20 & 84.00
& 50.61 & \second{48.10}
& 40.07 \\

SnapKV
& 27.43 & 30.93 & \second{38.05}
& 54.77 & \best{46.72} & \second{27.74}
& 29.84 & 22.67 & 24.07
& \second{38.50} & \second{91.41} & \second{41.98}
& \second{9.55} & \second{92.50}
& 53.25 & 46.77
& \second{42.26} \\

PyramidKV
& \second{27.45} & 30.24 & 36.83
& \best{55.62} & 44.43 & 26.81
& 28.75 & 22.81 & 23.91
& 35.50 & 91.26 & 41.21
& \best{10.55} & \best{95.00}
& \second{53.53} & 47.21
& 41.94 \\

\textbf{RippleKV}
& \best{28.79} & \second{33.53} & \best{40.74}
& \second{54.84} & \second{45.89} & \best{29.08}
& \second{30.15} & \second{22.93} & 24.00
& \second{38.50} & \second{91.41} & \best{42.33}
& \second{9.55} & \best{95.00}
& \best{53.79} & 46.91
& \best{42.97} \\
\midrule

\multicolumn{18}{c}{Qwen2.5-7B-Instruct, KV Cache Budget = 10\%} \\
\midrule
StreamingLLM
& \second{20.40} & \second{16.67} & 22.56
& 25.12 & 17.53 & 7.79
& 26.29 & \second{18.70} & 18.04
& \best{44.75} & 69.49 & 36.16
& 2.50 & 15.50
& 61.61 & \best{63.14}
& 29.14 \\

H$_2$O
& 13.39 & \best{20.30} & \best{27.54}
& 26.03 & \best{24.16} & 6.05
& \best{29.32} & \best{20.78} & \best{22.09}
& \second{43.00} & 81.36 & 35.92
& 4.00 & 38.50
& \second{63.06} & \second{62.69}
& 32.39 \\

SnapKV
& 19.98 & 14.04 & \second{25.28}
& \second{36.77} & 23.35 & \best{16.76}
& 26.30 & 18.38 & 18.87
& 42.00 & 87.47 & 40.14
& \second{6.00} & \second{46.00}
& \best{63.37} & 61.56
& \second{34.14} \\

PyramidKV
& 17.37 & 13.97 & 25.02
& 30.18 & 23.09 & 9.15
& 25.49 & 17.99 & 19.04
& 42.00 & \second{87.55} & \second{40.32}
& 3.50 & 35.83
& 62.96 & 61.04
& 32.16 \\

\textbf{RippleKV}
& \best{20.52} & 15.05 & 24.97
& \best{37.05} & \second{23.84} & \second{15.94}
& \second{26.72} & 18.48 & \second{19.31}
& 41.00 & \best{88.00} & \best{40.59}
& \best{7.00} & \best{47.50}
& 62.65 & 62.06
& \best{34.42} \\
\midrule

\multicolumn{18}{c}{Mistral-7B-Instruct-v0.3, KV Cache Budget = 10\%} \\
\midrule
StreamingLLM
& \best{18.65} & 13.10 & 23.29
& 31.23 & 20.23 & 14.02
& \second{25.35} & 19.74 & 18.69
& 35.00 & 52.14 & \second{32.20}
& 4.00 & 14.50
& 50.67 & 52.89
& 26.61 \\

H$_2$O
& 9.06 & \best{26.99} & 24.98
& 20.85 & \second{23.42} & 10.68
& \best{28.31} & \best{21.52} & \best{24.25}
& \best{47.50} & 85.45 & \best{39.29}
& \best{4.72} & 47.50
& 45.23 & 51.71
& 31.97 \\

SnapKV
& 17.69 & 13.61 & \best{30.62}
& 33.04 & 22.53 & \best{18.42}
& 25.32 & 20.24 & 20.54
& 38.00 & \second{87.90} & 28.51
& 3.78 & \second{53.50}
& \second{52.54} & \best{54.42}
& \second{32.54} \\

PyramidKV
& 16.84 & 13.58 & \second{30.15}
& \second{34.50} & 22.99 & 15.23
& 24.95 & 20.49 & \second{20.70}
& 38.50 & \best{88.21} & 29.12
& 3.96 & 49.50
& 52.43 & \second{54.38}
& 32.22 \\

\textbf{RippleKV}
& \second{18.37} & \second{13.73} & 29.72
& \best{35.49} & \best{23.78} & \second{17.69}
& 25.32 & \second{20.50} & 20.27
& \second{39.25} & 87.78 & 29.87
& \second{4.55} & \best{57.50}
& \best{52.85} & 54.32
& \best{33.19} \\
\bottomrule
\end{tabular}%
}

\caption{Performance comparison over LongBench datasets. The best result is highlighted in bold, and the second best is underlined.}
\label{tab:longbench-main}
\end{table*}

\subsection{Sensitivity Guided Layer Budget Allocation}
\label{sec:budget-allocation}

Given the sensitivity profile
$\mathbf{s}=\{s_\ell\}_{\ell\in\mathcal{A}}$,
RippleKV converts the response of each layer into a cache budget, where
$\mathcal{A}$ denotes the set of compressible layers. Raw KL responses may have different numerical scales across models and calibration sets. Directly using them as budget weights would therefore make the allocation sensitive to their absolute magnitude. We first normalize the sensitivity values as
\begin{equation}
\widehat{s}_\ell
=
\frac{s_\ell-s_{\min}}
{s_{\max}-s_{\min}},
\qquad
\ell\in\mathcal{A},
\label{eq:sensitivity-normalization}
\end{equation}
where
$s_{\min}=\min_{\ell\in\mathcal{A}}s_\ell$
and
$s_{\max}=\max_{\ell\in\mathcal{A}}s_\ell$.
When all layers have identical sensitivity, we set
$\widehat{s}_\ell=0$ for every layer, yielding uniform allocation. Otherwise,
$\widehat{s}_\ell\in[0,1]$ preserves the ordering of the sensitivity scores and makes the allocation invariant to positive affine transformations of the raw profile.

We then map each normalized score to a positive budget multiplier:
\begin{equation}
\widetilde{w}_\ell
=
\exp\left(
\log r\cdot\widehat{s}_\ell
\right)
=
r^{\widehat{s}_\ell},
\qquad
\ell\in\mathcal{A},
\label{eq:sensitivity-mapping}
\end{equation}
where $r\geq1$ controls the allocation disparity across layers. For a nonconstant sensitivity profile, the ratio between the largest and smallest multipliers before clipping is exactly $r$. Thus, $r=1$ recovers uniform allocation, while increasing $r$ assigns progressively more cache to sensitive layers. The mapping is monotonic, so a layer with a larger sensitivity score cannot receive a smaller initial multiplier than a less sensitive layer.

To avoid extreme allocations, RippleKV applies clipping followed by normalization:
\begin{equation}
\mathbf{w}
=
\operatorname{ClipNormalize}
\left(
\widetilde{\mathbf{w}},\gamma
\right),
\qquad
\frac{1}{|\mathcal{A}|}
\sum_{\ell\in\mathcal{A}}w_\ell
=
1,
\label{eq:weight-normalization}
\end{equation}
where $\gamma\geq1$ specifies the clipping interval
$[1/\gamma,\gamma]$. The multipliers are clipped using this interval and then rescaled to have mean one over the compressible layers. This step suppresses unusually large or small allocations while preserving the overall cache budget and the relative pattern induced by the sensitivity profile.

Finally, the budget assigned to layer $\ell$ is
\begin{equation}
b_\ell
=
\operatorname{round}
\left(
\frac{B}{|\mathcal{A}|}w_\ell
\right),
\qquad
\ell\in\mathcal{A},
\label{eq:layer-budget}
\end{equation}
where $B/|\mathcal{A}|$ is the average budget under uniform allocation. Layers with stronger perturbation responses therefore receive budgets above the uniform average, whereas less sensitive layers are compressed more aggressively. Since the sensitivity profile determines only the relative allocation, the same profile can be reused under different global cache budgets by changing $B$.

RippleKV changes only how the global budget is distributed across layers. Given $b_\ell$, the base compression method applies its original token scoring and selection rule to retain the required number of cached positions at layer $\ell$. In our implementation, we preserve the SnapKV scoring rule and replace only its uniform layer budgets with the sensitivity guided allocation. This isolates the effect of layer budget allocation from changes in token importance estimation and introduces no additional model evaluation during inference.

\begin{figure*}[t]
    \centering
    \includegraphics[width=0.98\textwidth]{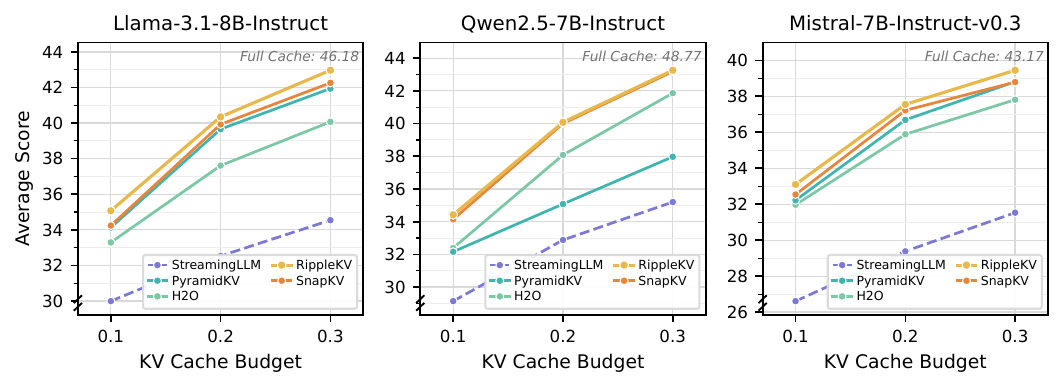}
    \caption{
    Average performance on LongBench under different KV cache budgets. 
    }
    \label{fig:budget_sensitivity}
\end{figure*}

\section{Experiments}
\label{sec:experiments}

\subsection{Experimental Settings}

\paragraph{Backbone Models.} We evaluate RippleKV on three widely used instruction-tuned LLMs: Llama-3.1-8B-Instruct~\citep{grattafiori2024llama}, Mistral-7B-Instruct-v0.3~\citep{DBLP:journals/corr/abs-2310-06825}, and Qwen2.5-7B-Instruct~\citep{hui2024qwen2}. These models represent different model families, enabling us to assess the generalizability of RippleKV across Transformer architectures.

\paragraph{Datasets.}We conduct experiments on LongBench~\citep{bai2024longbench}, a standard benchmark for long-context understanding. Following prior KV cache compression studies, we report results across six task categories: single-document QA, multi-document QA, summarization, few-shot learning, synthetic tasks, and code completion.

\paragraph{Baselines.} We use Full Cache, which preserves all KV states, as the reference setting. We compare RippleKV with StreamingLLM~\citep{xiao2024efficient}, which retains attention sinks and recent tokens; H$_2$O~\citep{zhang2023h2o}, which evicts tokens based on accumulated attention scores; SnapKV~\citep{li2024snapkv}, which selects tokens using a recent observation window; and PyramidKV~\citep{cai2024pyramidkv}, which allocates different cache budgets across layers. All compression methods use the same total cache budget.

\paragraph{Implementation Details.} All experiments are implemented using Hugging Face Transformers and PyTorch on NVIDIA A100 80GB GPUs. We follow the standard LongBench evaluation protocol with task-specific decoding settings. All compression methods are evaluated at cache retention ratios of 10\%, 20\%, and 30\%. Uniform baselines allocate the budget evenly across layers, whereas PyramidKV and RippleKV use non-uniform layer-wise allocation under the same total cache budget. We adopt the official or commonly used hyperparameters for all baselines.

\subsection{Main Results on LongBench}
\label{sec:main-results}



Table~\ref{tab:longbench-main} compares RippleKV with competing methods on LongBench. RippleKV achieves the highest average score across all five compressed settings. On Llama-3.1-8B-Instruct, it obtains average scores of 35.07, 40.35, and 42.97 under cache budgets of 10\%, 20\%, and 30\%, outperforming the strongest baselines by 0.83, 0.42, and 0.71 points, respectively. This consistent advantage demonstrates the robustness of RippleKV across different compression levels. Moreover, increasing the cache budget from 10\% to 30\% narrows its gap to Full Cache from 11.11 to 3.21 points. Figure~\ref{fig:budget_sensitivity} shows a similar trend across all three backbone models: RippleKV consistently retains the lead and steadily approaches Full Cache as more cache is preserved.

RippleKV also generalizes consistently across model families. Under the challenging 10\% cache budget, it achieves average scores of 34.42 on Qwen2.5-7B-Instruct and 33.19 on Mistral-7B-Instruct-v0.3, outperforming the strongest baselines by 0.28 and 0.65 points, respectively. Compared with the layer-aware PyramidKV, RippleKV improves the category-average Multi-Document QA score on Qwen2.5 from 20.81 to 25.61 and the Synthetic score on Mistral from 26.73 to 31.03. Although no method dominates every individual dataset, RippleKV consistently achieves the strongest aggregate performance across task categories. These findings demonstrate the effectiveness of allocating cache budgets according to end-to-end layer sensitivity rather than a fixed layer-wise allocation pattern.

\subsection{Efficiency Analysis}

\begin{table}[t]
\centering
\small
\setlength{\tabcolsep}{7pt}
\renewcommand{\arraystretch}{1.05}
\begin{tabular}{lcc}
\toprule
Method
& Latency (s)$\downarrow$
& Throughput (tok/s)$\uparrow$ \\
\midrule
\multicolumn{3}{c}{Context Length = 8K} \\
\midrule
Full Cache     & 7.58 & 37.38 \\
StreamingLLM   & 6.73 & 42.84 \\
SnapKV         & 6.84 & 42.37 \\
PyramidKV      & 6.88 & 42.09 \\
RippleKV       & 6.67 & 43.48 \\
\midrule
\multicolumn{3}{c}{Context Length = 128K} \\
\midrule
Full Cache     & 59.46 & 11.85 \\
StreamingLLM   & 43.35 & 42.23 \\
SnapKV         & 43.87 & 40.92 \\
PyramidKV      & 43.87 & 42.51 \\
RippleKV       & 43.02 & 42.99 \\
\bottomrule
\end{tabular}
\caption{Efficiency comparison at different context lengths.}
\label{tab:efficiency}
\end{table}

\begin{figure}[t]
    \centering
    \includegraphics[width=0.8\columnwidth]{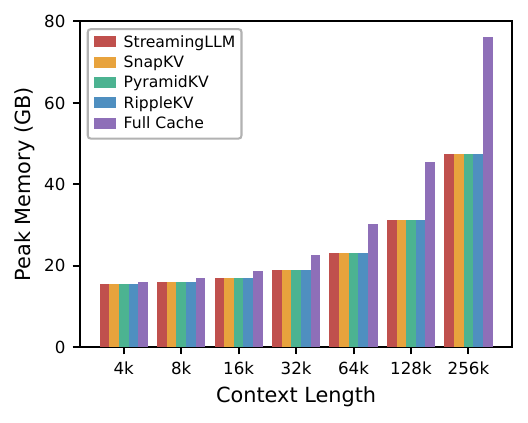}
    \caption{Peak memory across different context lengths.}
    \label{fig:peak-memory}
\end{figure}

Table~\ref{tab:efficiency} reports end-to-end inference latency and decoding throughput at representative context lengths of 8K and 128K, with additional results provided in the Appendix. At 8K, RippleKV achieves 6.67 seconds of latency and 43.48 tokens per second; at 128K, it records 43.02 seconds and 42.99 tokens per second, respectively. RippleKV thus matches or slightly outperforms existing KV cache compression methods in runtime efficiency, indicating that layer-wise budget allocation adds no measurable online overhead. The sensitivity profile is computed once during offline calibration, while inference only applies the precomputed budgets.

Figure~\ref{fig:peak-memory} further compares peak memory usage across context lengths from 4K to 256K. RippleKV closely matches the memory footprint of StreamingLLM, SnapKV, and PyramidKV, showing that redistributing cache capacity across layers does not increase the total cache budget. Compared with Full Cache, RippleKV reduces peak memory from 45.53 to 31.13 GiB at 128K and from 76.09 to 47.29 GiB at 256K, corresponding to reductions of 31.6\% and 37.8\%, respectively. These results show that RippleKV improves cache allocation while preserving the runtime and memory efficiency of existing KV cache compression methods.

\subsection{Ablation Study}
\label{sec:ablation}

\begin{table}[t]
\centering
\small
\setlength{\tabcolsep}{7pt}
\renewcommand{\arraystretch}{1.08}
\begin{tabular}{lccc}
\toprule
Category & w/o LBA & w/o DS & Full \\
\midrule
Single-Doc QA & 22.79 & 22.26 & \textbf{23.46} \\
Multi-Doc QA  & 29.52 & 29.35 & \textbf{30.89} \\
Summarization & 21.79 & 21.85 & \textbf{22.10} \\
Few-shot      & \textbf{55.45} & 54.87 & 55.37 \\
Synthetic     & 30.00 & 30.25 & \textbf{32.25} \\
Code          & 47.16 & 47.40 & \textbf{50.58} \\
\bottomrule
\end{tabular}
\caption{Ablation results across LongBench task categories.}
\label{tab:ablation}
\end{table}

We examine two key design choices in RippleKV: layer-wise budget allocation (LBA) and downstream sensitivity (DS). For \textit{w/o LBA}, we replace sensitivity guided allocation with a uniform cache budget across layers. For \textit{w/o DS}, we retain the same data driven allocation procedure and global cache budget, but replace the downstream signal, measured by the change in the final predictive distribution induced by perturbing the Value cache, with a layer local response score. This design isolates the contribution of downstream sensitivity measured at the model output from that of varying cache budgets across layers.

As shown in Table~\ref{tab:ablation}, the full model performs best on five of the six task categories. Removing LBA causes drops of 3.42 points on Code, 2.25 points on Synthetic, and 1.37 points on Multi-Document QA. Replacing downstream sensitivity with the layer-local signal yields corresponding drops of 3.18, 2.00, and 1.54 points. Although uniform allocation improves Few-shot Learning by 0.08 points, it underperforms the full model on all remaining categories. These results show that non-uniform layer allocation improves performance under a fixed cache budget, while downstream sensitivity provides a stronger allocation signal than layer-local responses.

\begin{table}[t]
\centering
\small
\setlength{\tabcolsep}{9pt}
\renewcommand{\arraystretch}{1.08}
\begin{tabular}{lccc}
\toprule
\multirow{2}{*}{Task Category}
& \multicolumn{3}{c}{Allocation Ratio $R$} \\
\cmidrule(lr){2-4}
& 1.25 & 1.50 & 1.75 \\
\midrule
Single-Doc QA  & 23.82 & 23.46 & 23.57 \\
Multi-Doc QA   & 29.74 & 30.89 & 30.33 \\
Summarization  & 21.81 & 22.10 & 22.01 \\
Few-shot       & 54.99 & 55.37 & 55.47 \\
Synthetic      & 31.50 & 32.25 & 31.75 \\
Code           & 50.29 & 50.58 & 50.31 \\
\midrule
Average        & 34.67 & 35.07 & 34.89 \\
\bottomrule
\end{tabular}
\caption{Sensitivity of RippleKV to the allocation ratio $R$ across LongBench task categories.}
\label{tab:ratio-sensitivity}
\end{table}

\subsection{Hyperparameter Sensitivity}
\label{sec:hyperparameter}

We study the sensitivity of RippleKV to the allocation ratio $R$, which controls the budget disparity between more and less sensitive layers. We vary $R$ over $\{1.25, 1.50, 1.75\}$ while keeping the global cache budget and all other settings fixed. As shown in Table~\ref{tab:ratio-sensitivity}, the corresponding average LongBench scores are 34.67, 35.07, and 34.89, with a maximum difference of only 0.41 points. Performance also remains stable across task categories, although the preferred value varies slightly by category. We therefore use $R=1.50$, which achieves the highest overall score, as the default setting. These results indicate that RippleKV is robust to moderate changes in the allocation ratio and does not rely on narrowly tuned hyperparameters.

\section{Conclusion}


In this work, we introduced RippleKV, a cross-layer KV cache allocation method guided by perturbation propagation. We showed that layer-wise compression damage is highly heterogeneous and does not follow a fixed depth pattern, limiting the reliability of allocation strategies based on structural priors. RippleKV instead measures how controlled Value cache perturbations affect the final predictive distribution and uses the resulting model dependent sensitivity profile to distribute a fixed global cache budget across layers. Experiments on LongBench across three model families and multiple cache budgets show that RippleKV consistently achieves the strongest average performance among the evaluated methods while maintaining comparable inference efficiency and memory usage. These results demonstrate that downstream output sensitivity provides an effective basis for cross-layer KV cache allocation.

\bibliography{aaai2027}

\end{document}